\documentclass[sigconf, review=false, 
screen=false, authorversion=true,
anonymous=false]{acmart}

\usepackage{epsfig}
\usepackage{graphicx}
\usepackage{amsmath}
\usepackage{amssymb}
\usepackage{bm}
\usepackage{url}            
\usepackage{booktabs}       
\usepackage{nicefrac}       
\usepackage{microtype}      
\usepackage{bbold}
\usepackage{algorithmic}
\usepackage{amsfonts}
\usepackage{mathtools,amssymb}
\usepackage{fixltx2e}
\usepackage{array}
\usepackage{setspace}
\usepackage{caption} 

\usepackage{multirow}
\usepackage{algorithm}

\newcommand{\eqn}[1]{Equation~#1}
\newcommand{\fig}[1]{Figure~#1}
\newcommand{\tab}[1]{Table~#1}

\newcommand{\etal}{\textit{et al}.}

\def\zero{\textcolor{white}{00}-\textcolor{white}{00}}

\AtBeginDocument{%
  \providecommand\BibTeX{{%
    \normalfont B\kern-0.5em{\scshape i\kern-0.25em b}\kern-0.8em\TeX}}}

\setcopyright{acmcopyright}
\copyrightyear{2020}
\acmYear{2020}
\acmDOI{10.1145/1122445.1122456}

\acmConference[MM '20]{MM '20: ACM Multimedia}{October 12--16, 2020}{Seattle, United States}
\acmBooktitle{MM '20: ACM Multimedia, October 12--16, 2020, Seattle, United States}
\acmPrice{15.00}
\acmISBN{978-1-4503-XXXX-X/18/06}

\acmSubmissionID{615}

\begin{document}

\title{Weakly Supervised 3D Object Detection from Point Clouds}

\author{Zengyi Qin}
\affiliation{\institution{Massachusetts Institute of Technology}}
\email{qinzy@mit.edu}

\author{Jinglu Wang}
\affiliation{\institution{Microsoft Research}}
\email{jinglu.wang@microsoft.com}

\author{Yan Lu}
\affiliation{\institution{Microsoft Research}}
\email{yanlu@microsoft.com}

\begin{abstract}
A crucial task in scene understanding is 3D object detection, which aims to detect and localize the 3D bounding boxes of objects belonging to specific classes. Existing 3D object detectors heavily rely on annotated 3D bounding boxes during training, while these annotations could be expensive to obtain and only accessible in limited scenarios. Weakly supervised learning is a promising approach to reducing the annotation requirement, but existing weakly supervised object detectors are mostly for 2D detection rather than 3D. In this work, we propose VS3D, a framework for weakly supervised 3D object detection from point clouds \textbf{without using any ground truth 3D bounding box for training}. First, we introduce an unsupervised 3D proposal module that generates object proposals by leveraging normalized point cloud densities. Second, we present a cross-modal knowledge distillation strategy, where a convolutional neural network learns to predict the final results from the 3D object proposals by querying a teacher network pretrained on image datasets. Comprehensive experiments on the challenging KITTI dataset demonstrate the superior performance of our VS3D in diverse evaluation settings. The source code and pretrained models are publicly available at \url{https://github.com/Zengyi-Qin/Weakly-Supervised-3D-Object-Detection}.
\end{abstract}



\maketitle

\section{Introduction}

\begin{figure*}
    \centering
    \includegraphics[width=\linewidth]{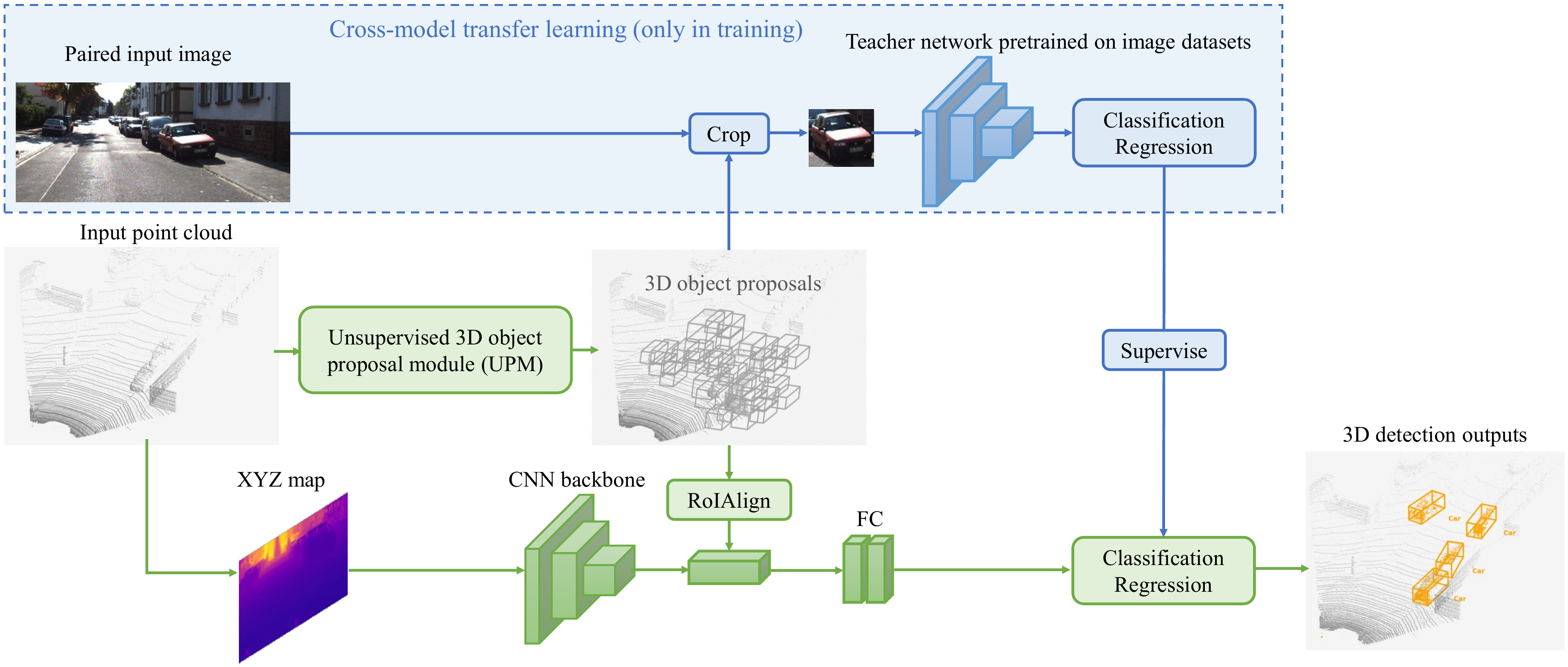}
    \caption{Overview of the proposed weakly supervised 3D object detection framework. The first key component is the unsupervised 3D object proposal module (UPM) that selects 3D anchors based on the normalized point cloud density. The second component is the cross-modal transfer learning module that transfers the knowledge, including object classification and rotation regression, from image datasets into the point cloud based 3D object detector. 
    }
    \label{fig:overview}
\end{figure*}

As an essential challenge in scene understanding, 3D object detection focuses on detecting and localizing the 3D bounding boxes of objects from input sensory data such as images and point clouds. Since point clouds offer a 3D geometric perception of the world, many 3D object detectors~\cite{chen2017multiview, hu2018joint, qi2017frustum, std2019yang} use point clouds as input data. These 3D object detectors transform unorganized point clouds into structured and compact 3D bounding box representations, and have been considered as pivotal components in mobile robots and augmented reality. However, training the 3D object detection algorithms would require human annotators to label a huge amount of amodal 3D bounding boxes in unorganized 3D point clouds. The annotation process could be labour-intensive and time-consuming. 

Most of the previous 3D detectors~\cite{chen2017multiview, mousavian20173dbox, qi2017frustum, qin2019monogr, roddick2018orthographic} are based on fully supervised learning and cannot adapt to scenarios where 3D labels are absent. Consequently, it is worthy of finding ways to achieve weakly supervised or even unsupervised learning of 3D object detection, which could reduce the requirement of training labels. Existing studies on weakly supervised learning of object detection mainly focus on 2D detection~\cite{inoue_2018_cvpr, Bilen16, kantorov2016}. However, 2D detection does not provide a 3D geometric understanding of the scene, which is crucial in various applications such as self-driving. Previous approaches~\cite{kowsari2016weighted, sun2019a} attempt to solve 3D object detection by leveraging non-parametric models without ground truth supervision, but they are not designed to provide the accurate 3D bounding boxes of objects. A recent work~\cite{tang2019transferable} focuses on semi-supervised 3D object detection, but it still assumes the existence of full 3D annotation for specific classes of objects.

In this work, we aim to develop a framework for weakly supervised 3D object detection from point clouds. 
We do not need ground truth 3D bounding boxes for training, but make full use of the commonly used data format, i.e., paired images and point clouds, for weak supervision.
Without ground truth, the \textbf{key challenges} of 3D object detection are \textit{1) how to generate 3D object proposals from unstructured point cloud} and \textit{2) how to classify and refine the proposals to finally predict 3D bounding boxes.} 
To solve the \textbf{first} challenge, we propose an \textbf{unsupervised 3D object proposal module (UPM)} that leverages the geometric nature of scan data to find regions with high object confidence. Point cloud density~\cite{chen20153dop} has been considered an indicator of the presence of an object. A volume containing an object could have a higher point cloud density, but the absolute density is also significantly affected by the distance to the scanner. Regions far away from the scanner are of low point cloud density even if they contain objects. To eliminate the interference of distance, we introduce the \textit{normalized point cloud density} that is more indicative of the presence of objects. 3D object proposals are generated by selecting the preset 3D anchors with high normalized point cloud density. However, the object proposals are class-agnostic, since we cannot distinguish the class of an object based on the normalized point cloud density. The rotation of an object is also ambiguous under the partial observation of the captured point clouds on its surface. Therefore, the pipeline should be able to classify the proposals into different object categories and regress their rotations, which reveal the second challenge.

To solve the \textbf{second} challenge, we propose a \textbf{cross-modal transfer learning} method, where the point cloud based detection network is regarded as a student and learns knowledge from an off-the-shelf teacher image recognition network pretrained on existing image datasets. The 3D object proposals produced by the UPM are projected onto the paired image and classified by the teacher network, then the student network mimic the behavior of the teacher during training. Using the teacher network as a media, we transfer the knowledge from the RGB domain to the targeted point cloud domain, which can save the annotation cost of 3D object detection on unlabeled datasets and facilitate the fast deployment of 3D object detectors in new scenarios. We notice that the teacher network is not always capable of supervising its student because of the gap between two different datasets, especially when the teacher is not confident of its own predictions. In light of this, we propose a rectification method that automatically strengthens confident supervisions and weakens uncertain ones. Thus the student learns more from reliable supervision signals while less from those unreliable. To validate the proposed approach and each of its components, we conduct comprehensive experiments on the challenging KITTI~\cite{geiger2012kitti} dataset. Promising performance is shown under diverse evaluation metrics. Our method demonstrates over $50\%$ improvement in average precision compared to previous weakly supervised object detectors. In summary, our contributions are three-fold:

\begin{itemize}
\item An unsupervised 3D object proposal module (UPM) that selects and aligns anchors using the proposed normalized point cloud density and geometry priors. 
\item An effective approach to transferring knowledge from 2D images to 3D domain, which makes it possible to train 3D object detectors on unlabeled point clouds.
\item A pioneering framework for weakly supervised learning of 3D object detection from point clouds, which is examined through comprehensive experiments and demonstrates superior performance in diverse evaluation settings.
\end{itemize}

\section{Related Work}

\begin{figure*}[t]
	\centering
	\includegraphics[width=\linewidth]{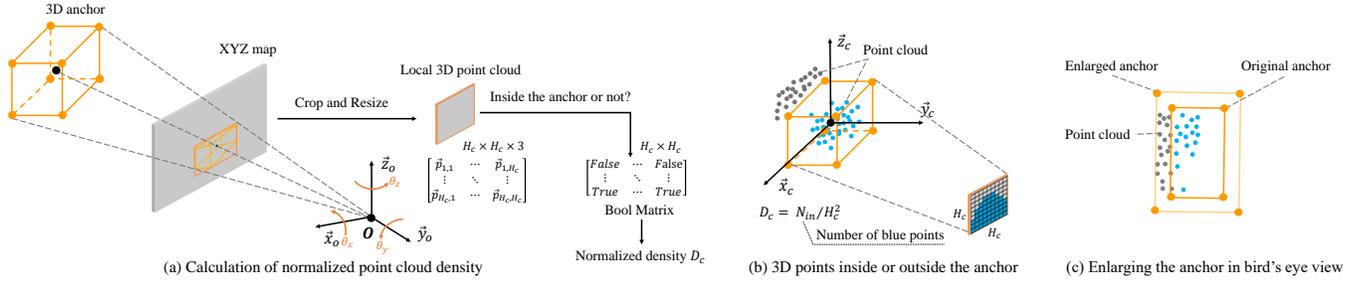}
	\caption{Normalized point cloud density. The point cloud density inside a volume is influenced by two factors that are 1) whether the volume contains an object and 2) the distance of the volume to the sensor. The density increases when an object is present but decreases as the distance grows. Our normalization strategy eliminates the influence of distance. (a) The preset 3D anchor is projected to the XYZ map, where its projection is cropped out and scaled to a square patch with fixed size $H_c \times H_c$ that is \textbf{distance-irrelevant}. The square patch represents $H_c^2$ 3D points, where one pixel corresponds to one point. (b) Among the 3D points, $N_{in}$ points are inside the 3D anchor. The normalized point cloud density $D_c$ is calculated as $N_{in}/H_c^2$. (c) The grey and blue points are on the same object. If the original anchor fails to bound the object and only contains a part of it, the enlarged version would contain more points, i.e., the grey ones.
	}
	\label{fig:pc_density}
\end{figure*}

\subsection{3D object detection}
The objective of 3D detection is detecting objects of interest and localizing their amodal 3D bounding boxes. Existing approaches are based on full supervision, assuming that the accurate 3D ground truth is provided in the dataset. MonoGRNet~\cite{qin2019monogr} proposes predicting the 3D location by first estimating instance-level depth and the projected 3D center. TLNet~\cite{qin2019tlnet} triangulates the objects using stereo images. MV3D~\cite{chen2017multiview} introduces bird's eye view (BEV) representation of point cloud data to construct region proposal network~\cite{ren2017faster}. F-PointNet~\cite{qi2017frustum} detects object on image to reduce search space in LiDAR point cloud. VoxelNet~\cite{zhou2018voxelnet} groups 3D point cloud into each voxel where a fix-length feature vector is extracted. AVOD~\cite{hu2018joint} aggregates the image view and bird's eye view (BEV) to produce high-quality object proposals. STD~\cite{std2019yang} transforms the point cloud features from sparse to dense. The state-of-the-art performance of these models are established on sufficient training labels. 

\subsection{Weakly supervised object detection}
Weakly supervised object detection assumes that the instance-level bounding box annotations are not provided by the training set, and the supervision can come from image-level annotations. Cho \etal~\cite{cho2015unsupervised} proposes to discover dominant objects and localize their 2D bounding boxes via a part-based region matching approach in a fully unsupervised fashion. With image-level labels, Han \etal~\cite{han2015object} propose learning 2D localization by iterative training. Sangineto~\etal~\cite{sangineto2019self} select object proposals that are more confident in training. WSDDN~\cite{Bilen16} modifies image classification networks to predict at the region level for object proposal selection and classification. OICR~\cite{tang2017oicr} and PCL~\cite{tang2018pcl} utilize online instance classification refinement and proposal cluster learning to improve the detection performance. OIM~\cite{lin2020oim} leverages instance graphs to mine all possible instances using image-level annotations. Although there are studies on 2D object detection in unsupervised or weakly supervised settings, 3D detection without ground truth supervision is much less explored. The work of~\cite{kowsari2016weighted} attempts to solve the unsupervised 3D object detection by performing weighted clustering on point clouds, but is not designed to predict the amodal 3D bounding boxes. Reasoning at region level, object detection is already a nontrivial task on 2D images. Learning 3D detection without full supervision is more challenging and will be explored in this paper.

\subsection{Cross-modal transfer learning}
Knowledge distillation~\cite{hinton2015distilling} is widely used for transferring supervision cross moralities, such as \cite{gupta2016cross,abdulnabi2018multimodal,garcia2018modality}. Huang \etal~\cite{huang2017like} minimize the neural activation distribution of the teacher and the student network. Sungsoo \etal~\cite{ahn2019variational} propose maximizing the high-level mutual knowledge between the teacher and its student instead of matching their activation or other handcrafted features.
Gupta \etal \cite{gupta2016cross} propose to distill semantic knowledge from labeled RGB images to unlabeled depth for 2D recognition tasks, e.g., 2D detection and segmentation, while the 3D geometric information in depth data is not fully utilized. Instead, our method does not require labels of the RGB images on the targeted dataset, and also explores the depth information for object proposal in 3D space.

\section{Approach}

\subsection{Overview}

The objective is to detect and localize amodal 3D bounding boxes of objects from the input point clouds. Unlike existing 3D detectors, we do not rely on the ground truth 3D bounding boxes on the targeted datasets during training. As is shown in \fig{\ref{fig:overview}}, the detection pipeline consists of two stages, 1) an unsupervised 3D object proposal model (UPM) and 2) a cross-modal transfer learning method. The first UPM stage outputs object proposals indicating the regions potentially containing the objects from point clouds. The second transfer learning stage classifies and refines the proposals to produce the final predictions by leveraging a teacher model pretrained on image datasets. LiDAR scanners are not a necessity in providing the input point clouds, which could also be obtained from a monocular image~\cite{Weng2019pseudolidarmono} or a pair of stereo images~\cite{wang2019pseudo}. It is assumed that each frame of point clouds has a paired image in the training set, but this is not required in testing where only the point clouds are needed. This assumption is satisfied by most datasets.

\subsection{Unsupervised 3D object proposal module}
\label{sect:norm_density}
3D object proposals are defined as volumes potentially containing objects. We first preset 3D anchors and then select anchors with high object confidence as object proposals. The preset anchors are placed at an interval of 0.2m on the ground plane spanning $\text{[0, 70m]} \times \text{[-35m, 35m]}$. Without ground truth supervision, it is not feasible to directly train a model to select anchors as the proposals from raw point clouds. Therefore, we explore the geometric nature of point clouds and leverage our prior knowledge to find potential regions with objects. The point cloud density inside a volume can indicate whether an object is contained in that volume. A high density represents a high objectiveness confidence. However, the point density is also significantly influenced by the distance to sensor. Distant points are much sparsely distributed than nearby points. Hence, we introduce a distance-invariant point density measurement, \textbf{normalized point cloud density} $D_c$, for effectively selecting potential candidate anchors.

\subsubsection{Normalized point cloud density.} 

We project the 3D point cloud to the front view to obtain the pixel-wise XYZ map, where each pixel has three channels indicating the 3D coordinates. The empty pixels are filled by inpainting~\cite{bertalmio2000inpaint}. By projecting the 3D anchor to the front view, we get a 2D box that bounds the projection, as illustrated in \fig{\ref{fig:pc_density}} (a). We crop the patch of XYZ map inside the bounding box and resize it into $H_c\times H_c$ by interpolation, after which we obtain $H_c^2$ 3D points, where each point is denoted as $\vec{p}_{i, j}$ as is shown in \fig{\ref{fig:pc_density}}~(a). It is noteworthy that so far the resized front-view patch of each anchor consists the \emph{same} number of 3D points regardless of the distance-relevant sparsity of point cloud. Some of these points are inside the anchors, denoted as \emph{True} in the boolean matrix in \fig{\ref{fig:pc_density}}~(a), while those outside the anchor are \emph{False}. Among those points, if there are $N_{in}$ points inside the 3D anchor, its point cloud density $D_c$ can be expressed as $N_{in} / H_c^2$. $N_{in}$ does not include the points on the ground planes, which are identified by RANSAC~\cite{fischler1981ransac} based plane fitting. $D_c$ is not influenced by the distance because we interpolate the front-view patch to the same size. If a targeted object is contained in an anchor, $D_c$ of this anchor should be above a certain threshold $\delta$.

The normalized point cloud density requires calculating $N_{in}$, i.e., identifying how many points among those $H_c^2$ points are inside the 3D anchor. \fig{\ref{fig:pc_density}}~(b) illustrates how to distinguish whether point $\vec{p}$ is inside the anchor. By transforming the 3D point $\vec{p}$ from the camera coordinate system to the local anchor coordinate system, we obtain $(q_x, q_y, q_z) = ((\vec{p} - \vec{c}) \cdot \vec{x}_c, (\vec{p} - \vec{c}) \cdot \vec{y}_c, (\vec{p} - \vec{c}) \cdot \vec{z}_c)$, where $\vec{c}$ denotes the translation from the origin to the anchor center, $\vec{x}_c$, $\vec{y}_c$ and $\vec{z}_c$ are the axis aligned to anchor dimensions. Denote half of the anchor size along each axis as $l_x$, $l_y$ and $l_z$. The element $b_{i, j}$ in the boolean matrix of \fig{\ref{fig:pc_density}}~(a) is determined by:
\begin{align}
    b_{i, j} = (|q_{i, j, x}| < l_x) \wedge (|q_{i, j, y}| < l_y) \wedge (|q_{i, j, z}| < l_z)
    \label{eq:boolmat}
\end{align}
where $q_{i,j,x}$, $q_{i,j,y}$, $q_{i,j,z}$ are the transformed XYZ coordinates of points at index of $(i,j)$ in the patch.  
Please note that this process can be done in parallel in GPU implementation for all the points in all the front-view patches. Different from the forward pass in a convolutional neural work, the whole computation does not involve extensive floating point operations, demanding only a little computational resource and time cost. Learning is not required here, which further enhances the efficiency of the pipeline.

\subsubsection{Anchor selection and alignment} 

The anchors with $D_c<\delta$ are considered negative examples and are efficiently removed. The 3D object proposals are selected from the remaining anchors. We enlarge each anchor by $1+\epsilon$ times as is in \tab{\ref{fig:pc_density}}~(c). If the enlarged anchor does not contain extra points excluding the points in the original anchor, the original anchor is selected as one of the object proposals. This ensures that the anchor bounds the whole object rather than only a part of it, otherwise the enlarged anchor could contain extra points. $\epsilon$ is small enough to avoid the enlarged anchor containing point clouds from the neighbouring objects.

The selected anchors may not well aligned with the targeted object. We observe that some points should be close to the anchor's rectangular surfaces if the anchor fits to the target. In light of this, we shift the anchor in a small range so that it is better aligned with the points it contains. Specifically, for the axis of $\vec{x}_c$, we find the point $\vec{p}_{i, j}$ that is inside the anchor and has the greatest projected length $|q_{i, j, x}|$, then we move the anchor along $\vec{x}_c$ to align the closest surface with $\vec{p}_{i, j}$. The alignment along $\vec{y}_c$ and $\vec{z}_c$ are in the same way. Similar to the calculation in \eqn{\ref{eq:boolmat}}, this anchor alignment is done in parallel for all anchors efficiently.

\subsection{Image to point cloud knowledge transfer}
The object proposals produced by UPM are not the final detection outputs. It is observed that some selected anchors will contain objects not belonging to the targeted category. For instance, an anchor of car class may bound points of trees, possessing a a higher point cloud density than the selection threshold. As no ground truth is provided, it is difficult to recognize object category from point clouds. In addition, since the object rotations are hard to determine based on partially observed point cloud, object proposals from UPM are of the same rotations as pre-defined. One possible way to enhance the recognition ability is introducing knowledge from another domain. There is a large quantity of RGB data~\cite{xiang_wacv14} that are already labeled. We expect to transfer the knowledge from the RGB domain to the point cloud domain. To this end, we propose a cross-modal transfer learning method, where the point cloud based 3D detector is regarded as student and learns knowledge from a teacher network pre-trained on large-scale image recognition datasets.

\subsubsection{Image based teacher network}
\label{sect:pretrained_teacher}
The teacher is an image recognition and view point regression network employing the VGG16~\cite{simonyan2014very} architecture and is pre-trained on the ImageNet~\cite{imagenet15} and PASCAL VOC~\cite{pascalvoc} with image-level classification labels and view point labels provided by \cite{xiang_wacv14}. This is consistent with previous work on weakly supervised object detection~\cite{tang2018pcl, tang2017oicr, Bilen16, lin2020oim} where annotated bounding boxes are assumed absent during training. Taking an image with no more than one object as input, the teacher network classifies the image as background or a class of objects, and at the same time regresses the object view point as its rotation. The view point regression is considered a multi-bin classification problem, where we predict the probabilities of 16 angle bins divided from a unit circle. The rotation output is the expectation value of the angles of all bins. Neither of the datasets has overlaps with KITTI dataset where we evaluate our approach in the experiment. The teacher is used as an off-the-shelf model in training of the 3D object detection model, shown as the blue branch of \fig{\ref{fig:overview}}.

\subsubsection{Point cloud based student network}
\label{sect:student}

The student represents the second stage of a point cloud based 3D object detector, consisting of a VGG16~\cite{matthew2014vgg} backbone, a RoIAlign~\cite{he2017mrcn} layer and the fully connected layers, as is illustrated in the green branch in \fig{\ref{fig:overview}}. The input point clouds are converted to a front-view XYZ map before fed into the backbone similar as the work~\cite{wu2016SqueezeDet}. Using the image paired with the point clouds, we can distill the recognition knowledge from the teacher network to the student. More specifically, we project each object proposal produced by UPM to both the RGB image and the front-view XYZ map. Then we crop out the projection on the image and recognize the object proposal using the teacher network. At the same time, we use RoIAlign~\cite{he2017mrcn} to extract encoded features of each proposal from the student backbone and feed the features to fully connected layers to predict the object class and rotation. During training, each object proposal has two predictions from the teacher and the student respectively. The student learns to imitate the confidence of the teacher using the rectified cross-entropy loss as described in the followings.

\begin{figure}
    \centering
    \includegraphics[width=\linewidth]{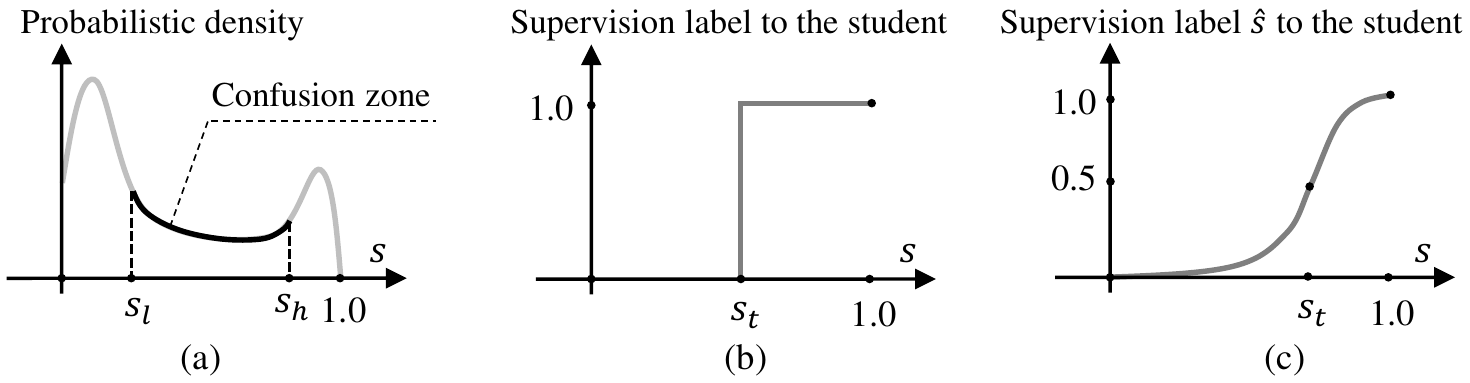}
    \caption{Supervision rectification. Given an object proposal to be classified, if the confidence of the teacher network is within the confusion zone, the loss value will be masked out.}
    \label{fig:confusion_alleviation}
\end{figure}

When distilling the capability from an off-the-shelf teacher into a student of a different dataset, there could inevitably occur problems. First, the teacher may not be confident of some of its own outputs. Such confusing outputs are supposed not to teach the student. More specifically, as is shown in \fig{\ref{fig:confusion_alleviation}}~(a), when the classification score $s$ from the teacher is within the confusion zone, $s$ will not be used to train the student branch. Only when the teacher is confident enough will the student learn from it. Second, all the predictions above the confusion zone should not be treated equally as is in \fig{\ref{fig:confusion_alleviation}}~(b), where the supervision label to the student is a binary value. In \fig{\ref{fig:confusion_alleviation}}~(b), if the score predicted by teacher network is above $s_t$, the supervision label to the student network will be $1.0$, indicating an absolute positivity. However, this makes all positive labels indistinguishable. In light of this, we propose supervision rectification, where we rectify the teacher's output $s$ to produce a soft label $\hat{s}$ as is illustrated in \fig{\ref{fig:confusion_alleviation}}~(c). The rectified cross entropy loss is formulated: 
\begin{equation}
\resizebox{.9\hsize}{!}{$
\mathcal{L}_{r} = -\left[\hat{s}\log(\Tilde{s}) + (1-\hat{s})\log(1-\tilde{s})\right]\cdot\mathbb{1}(s \not\in [s_l, s_h]) 
$}
\label{eq:spv_rect} 
\end{equation}
where $\tilde{s}$ is the prediction of the green branch to be supervised. $\mathbb{1}(\cdot)$ refers to the indicator function whose value is 1 iff $s$ from the teacher is not in the confusion zone shown in \fig{\ref{fig:confusion_alleviation}}~(a). The relationship between $\hat{s}$ and $s$ is given by the following function:
\begin{align}
    \hat{s} = \frac{1+e^{(s_t-1)k}}{1+e^{(s_t-s)k}}
    \label{eq:exp_rect}
\end{align}
See \fig{\ref{fig:confusion_alleviation}}~(c) for the curve of \eqn{\ref{eq:exp_rect}}, where $s_t$ is a soft threshold and $k$ controls slope. The rectified label takes into account the confidence of teacher network. Higher confidence leads to stronger positivity in the supervision labels provided to the student network.

\section{Experiment}

\begin{table*}[t]
\centering
\caption{Object detection 2D recall on the public KITTI validation set comparing with weakly supervised methods.}
\resizebox{\textwidth}{!}{
\setlength{\tabcolsep}{2mm}{
\begin{tabular}{c c c c c c c c c c c c c c}
 \toprule
 \multirow{2}{*}{Method} & \multirow{2}{*}{Input} &&  \multicolumn{3}{c}{Recall (IoU = 0.3)} && \multicolumn{3}{c}{Recall (IoU = 0.5)} && \multicolumn{3}{c}{Recall (IoU = 0.7)} \\ \cmidrule{4-14}
                         &     && Easy & Moderate & Hard && Easy & Moderate & Hard && Easy & Moderate & Hard \\\midrule
   PCL~\cite{tang2018pcl}         & Mono   &&49.08 &32.68 &29.76 &&24.04 &15.23 &13.41 &&6.461 &3.948 &3.356  \\
   OICR~\cite{tang2017oicr}        & Mono   &&56.42 &43.96 &39.55 &&25.63 &18.71 &16.59 &&6.191 &4.547 &3.768  \\
   MELM~\cite{wan2018min}        & Mono   &&64.52 &58.37 &53.56 &&27.70 &24.81 &21.93 &&7.234 &6.275 &5.099  \\ \midrule

   VS3D & Mono           &&94.09 &87.26 &76.41 &&90.12 &80.76 &67.45 &&63.83 &51.28 &40.38 \\
   VS3D & Stereo         &&94.30 &87.86 &76.84 &&90.99 &82.53 &69.13 &&64.61 &53.29 &41.70\\
   VS3D & LiDAR          &&90.92 &83.57 &73.71 &&86.60 &77.02 &65.11 &&60.31 &48.26 &38.51\\ \midrule
   VS3D & Mono + LiDAR   &&94.48 &88.27 &77.62 &&92.65 &82.58 &69.33 &&65.10 &53.07 &42.04 \\
   VS3D & Stereo + LiDAR &&95.00 &88.63 &78.15 &&91.08 &83.03 &70.11 &&65.58 &53.93 &42.74 \\
\bottomrule
\end{tabular}}}
\label{tab:recall}
\end{table*}

\begin{table*}[t]
\centering
\caption{Object detection average precision~(AP) on KITTI validation set comparing with weakly supervised methods.}
\resizebox{\textwidth}{!}{
\setlength{\tabcolsep}{1.8mm}{
\begin{tabular}{c c c c c c c c c c }
 \toprule
 \multirow{2}{*}{Method} & \multirow{2}{*}{Input} &&  \multicolumn{3}{c}{AP\textsubscript{2D}~/~AP\textsubscript{3D} (IoU = 0.3)} && \multicolumn{3}{c}{AP\textsubscript{2D}~/~AP\textsubscript{3D} (IoU = 0.5)} \\ \cmidrule{4-10}
                         &     && Easy & Moderate & Hard && Easy & Moderate & Hard \\\midrule
   PCL~\cite{tang2018pcl}         & Mono   &&5.916 / \zero &4.687 / \zero &3.765 / \zero &&1.878 / \zero &1.058 / \zero &0.935 / \zero \\
   OICR~\cite{tang2017oicr}        & Mono   &&13.50 / \zero &8.604 / \zero &8.045 / \zero &&6.481 / \zero &2.933 / \zero &3.270 / \zero \\
   MELM~\cite{wan2018min}        & Mono   &&8.054 / \zero &7.282 / \zero &6.882 / \zero &&2.796 / \zero &1.486 / \zero &1.476 / \zero \\ \midrule

   VS3D & Mono            &&77.73 / 55.90 &73.82 / 48.83 &65.71 / 40.92 &&76.93 / 31.35 &71.84 / 23.92 &59.39 / 19.34 \\
   VS3D & Stereo          &&79.04 / 70.72 &75.90 / 63.78 &67.55 / 52.03 &&79.03 / 40.98 &72.71 / 34.09 &59.77 / 27.65 \\
   VS3D & LiDAR           &&78.64 / 65.96 &74.41 / 59.76 &66.24 / 49.78 &&74.54 / 40.32 &66.71 / 37.36 &57.55 / 31.09 \\ \midrule
   VS3D & Mono + LiDAR    &&82.46 / 69.75 &78.84 / 63.47 &69.36 / 52.76 &&81.60 / 41.83 &72.43 / 39.22 &64.31 / 32.73 \\
   VS3D & Stereo + LiDAR  &&82.84 / 70.09 &78.99 / 65.25 &69.83 / 55.77 &&81.95 / 42.43 &73.21 / 41.58 &64.34 / 32.74 \\
\bottomrule
\end{tabular}}}
\label{tab:averageprecision}
\end{table*}

\begin{figure*}
    \centering
    \includegraphics[width=\linewidth]{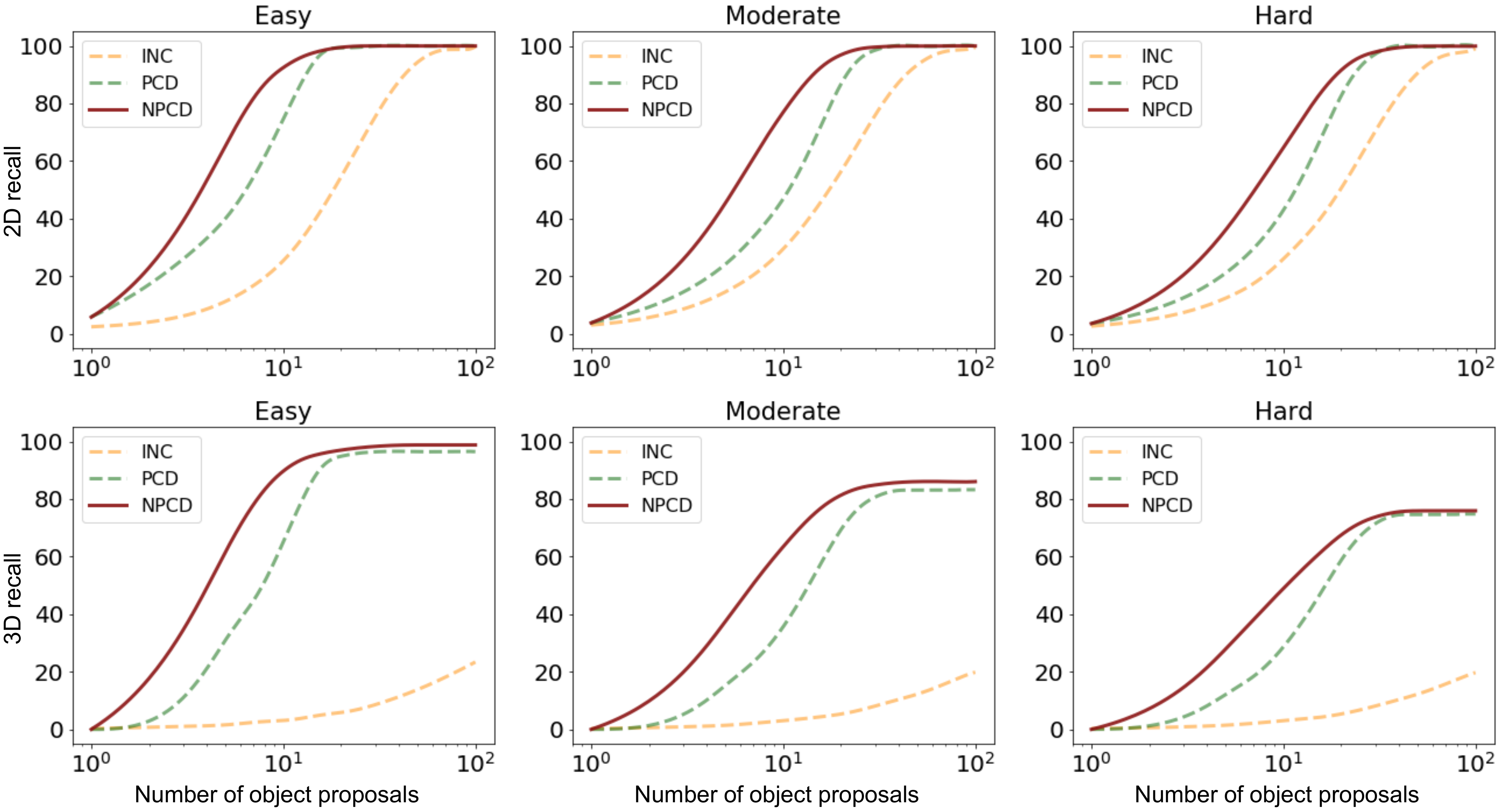}
    \caption{Bounding box recall using different unsupervised 3D object proposal methods. The first and second rows indicate 2D and 3D recall respectively. Results are obtained on KITTI validation set under the IoU threshold of 0.1. NPCD represents the proposed approach based on normalized point cloud density. PCD denotes point cloud density. INC is an inclusive method where all the anchors are kept. It is shown that NPCD uses fewer object proposals to reach a higher 3D recall rate.}
    \label{fig:recall_curve}
\end{figure*}

\begin{table}[t]
\centering
\caption{The gap between the weakly supervised VS3D and fully supervised methods in 3D object detection.}

\resizebox{0.46\textwidth}{!}{
\setlength{\tabcolsep}{1mm}{
\begin{tabular}{c c c c c c }
 \toprule
 \multirow{2}{*}{Method} &  \multirow{2}{*}{Input} &  \multirow{2}{*}{Sup. type} &  \multicolumn{3}{c}{AP\textsubscript{3D} (IoU = 0.3)}\\ \cmidrule{4-6}
         &     &     & Easy & Moderate & Hard \\\midrule
Deep3DBox~\cite{mousavian20173dbox}  & Mono   & Full   & 54.30 & 43.42 & 36.57 \\
MonoGRNet~\cite{qin2019monogr}      & Mono    & Full   &72.17 &59.57 &46.08 \\
VoxelNet~\cite{zhou2018voxelnet}       & LiDAR   & Full   &89.32 &85.81 &78.85 \\ \midrule
VS3D           & Mono    & Weak  &55.90 &48.83 &40.92  \\
VS3D           & LiDAR   & Weak  &65.96 &59.76 &49.78 \\ 
\bottomrule
\end{tabular}}}
\label{tab:gapweakfull}
\end{table}

\begin{table}[t]
\centering
\caption{Using our unsupervised object proposal module to improve existing weakly supervised object detectors.}

\resizebox{0.46\textwidth}{!}{
\setlength{\tabcolsep}{1mm}{
\begin{tabular}{c c c c c}
 \toprule
 \multirow{2}{*}{Method} &  \multirow{2}{*}{Input} &  \multicolumn{3}{c}{AP\textsubscript{2D} (IoU = 0.3)}\\ \cmidrule{3-5}
             &          & Easy & Moderate & Hard \\\midrule
   PCL~\cite{tang2018pcl}       & Mono     &5.916 &4.687 &3.765 \\
   OICR~\cite{tang2017oicr}      & Mono     &13.50 &8.604 &8.045 \\
   MELM~\cite{wan2018min}      & Mono     &8.054 &7.282 &6.882 \\ \midrule
   
   PCL~\cite{tang2018pcl} + UPM     & Mono + LiDAR      &16.69 &16.53 &13.90 \\
   OICR~\cite{tang2017oicr} + UPM    & Mono + LiDAR      &30.87 &27.97 &26.98 \\
   MELM~\cite{wan2018min} + UPM    & Mono + LiDAR      &18.41 &17.65 &15.55 \\

\bottomrule
\end{tabular}}}
\label{tab:apwithupm}
\end{table}

\begin{figure*}[t]
    \centering
    \includegraphics[width=\linewidth]{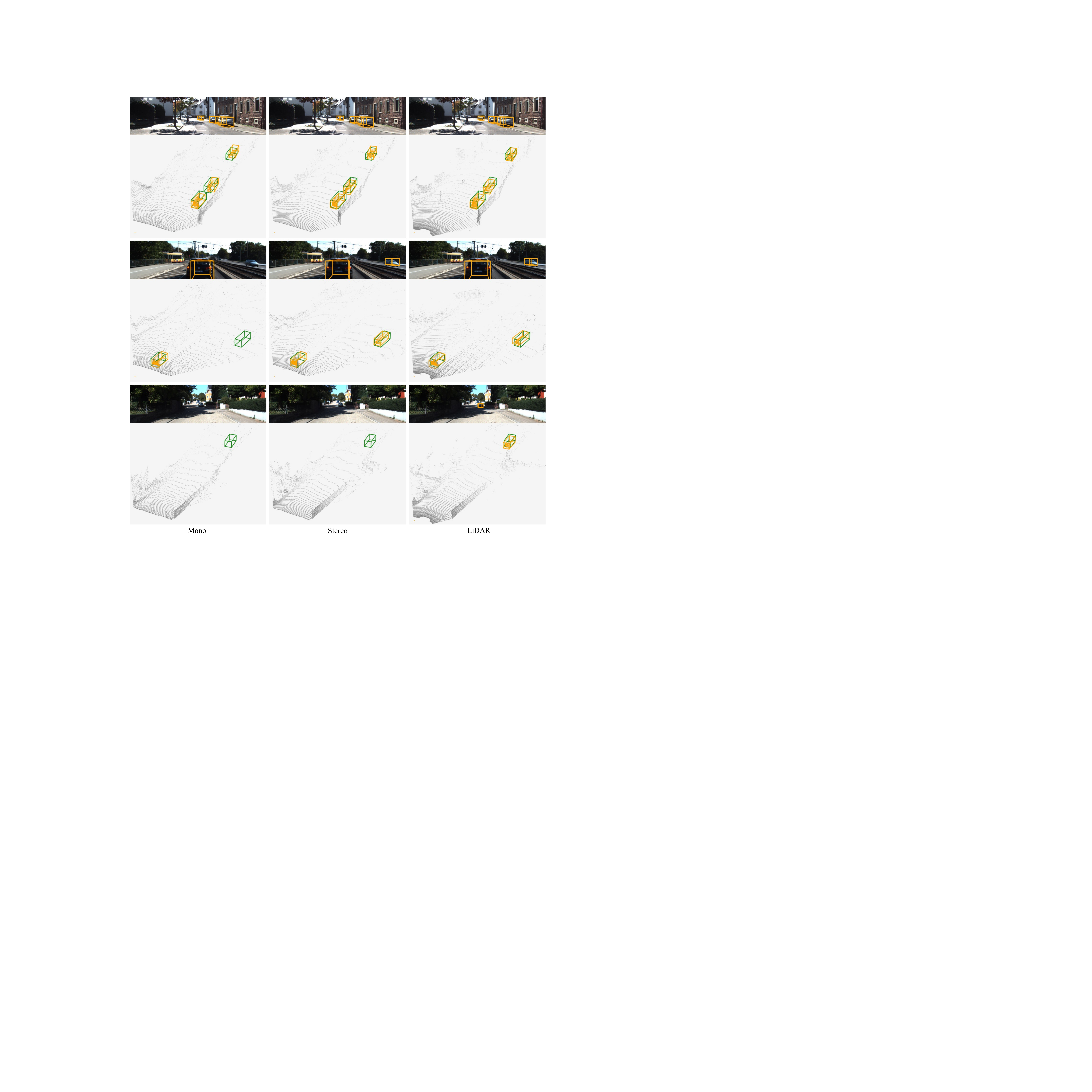}
    \caption{Qualitative results of VS3D on KITTI validation set. Predictions are shown in orange while the ground truths are in green. LiDAR based VS3D is more robust in detecting distant objects in the shadow, as is shown in the third row.}
    \label{fig:qualitative}
\end{figure*}

The evaluation is done on the KITTI~\cite{geiger2012kitti} dataset, where the publicly available training and validation set are split following~\cite{chen2016monocular, chen20153dop, chen2017multiview} Both of the splits contain half of the whole training set and has no overlap in terms of the video sequences where the frames come from. 

\noindent\textbf{Metrics.} Various metrics including recall rate and average precision (AP) with different intersection of union (IoU) thresholds are utilized to provide a thorough evaluation of the proposed method. If the IoU between a predicted bounding box and a ground truth bounding box is no less than a given threshold, then the prediction is considered correct and the ground truth is recalled. Recall rate measures the proportion of ground truth that is recalled. AP calculates the precision average across different recall thresholds. These metrics have been widely adopted in previous works~\cite{pascalvoc, tang2018pcl, tang2017oicr, qin2019monogr, qin2019tlnet}. 

There are three fundamental questions that we aim to answer: 1) How is the quantitative performance of the proposed detection framework and its comparison to existing methods? 2) How does the performance change with respect to different types of input signals including monocular images, stereo images and LiDAR scans? 3) How important is the unsupervised 3D object proposal module to the whole framework?

\subsection{Implementation.} 

The framework is built on Tensorflow \cite{abadi2016tensorflow}. Both the teacher and the student network employ the VGG16~\cite{matthew2014vgg} backbone. In the teacher network, we scale the input images to a fixed size $64\times64$ to introduce scale invariance. In the student network, RoI features are from the last conv layer of the backbone. The features are resized to $7\times7$ and passed to three consecutive fully connected layers with 1024 and 512 hidden units, where the last layer outputs the classification probability and the multi-bin probability for rotation prediction. There are 16 bins in total. In supervision rectification, we consider object proposals with $s>s_h$ as positive examples and $s<s_l$ as negative examples. In training, a mini-batch has 1024 positive and 1024 negative examples. The top 512 object proposals are kept in inference. For the hyperparameters, we choose $H_c=32, \delta=0.5, \epsilon=0.2, s_t=0.6$, $s_l=0.4$ and $s_h=0.6$ by grid search. The whole network is trained using Adam~\cite{kin2015adam} optimizer for 40 epochs with a constant learning rate of $10^{-4}$. L2 regularization is applied to model parameters at a decay weight of $5\times10^{-5}$. 

\noindent\textbf{Input type.} A frame of input point clouds could be obtained from three sources including a monocular image, a pair of stereo images and LiDAR scans. For the monocular image, we feed it to DORN~\cite{fu2018ordinal} to predict the pixel-level depths then convert the depths to 3D point clouds. For the stereo images, we feed them to PSMNet~\cite{chang2018pyramid} to produce the depths that are transformed to 3D point clouds. For LiDAR, point clouds are directly accessible. Corresponding to the data types, we train and evaluate three versions of our framework. At test time, a single forward pass from input point clouds to the output 3D bounding boxes takes 44ms on a Tesla P40 GPU, demanding 9.39 billion floating point operations~(FLOPS). The fast anchor selection stage involves 9.56 million FLOPS that is only 0.1\% of the total computational cost. Most of the resource is consumed by the backbone network, meaning that the efficiency can be improved when a lighter backbone is employed.

\subsection{Weakly supervised object detection}

Three state-of-the-art weakly supervised detection methods~\cite{tang2018pcl, wan2018min, tang2017oicr} are compared. PCL~\cite{tang2018pcl} iteratively learns refined instance classifiers by clustering the object proposals. OICR~\cite{tang2017oicr} adds online instance classification refinement to a basic multiple instance learning network. MELM~\cite{wan2018min} builds a min-entropy latent model to measure the randomness of object localization and guide the discovery of potential objects. The original papers do not provide results on the KITTI~\cite{geiger2012kitti} dataset, but the authors have made their code publicly available. Strictly following the guidelines of the code base, all the models are retrained and evaluated on the KITTI~\cite{geiger2012kitti} dataset. Since these methods cannot predict 3D bounding boxes, the comparison would be mainly in 2D domain. Three versions of our VS3D are also evaluated, corresponding to monocular, stereo and LiDAR inputs.

\tab{\ref{tab:recall}} presents the recall under different IoU thresholds using the top 10 predictions per frame. It is shown that our method outperforms MELM~\cite{wan2018min} by $20\%$ to $50\%$, which could be interpreted as a huge margin in terms of recall rate. It can be observed that the margin grows as the evaluation metric becomes stricter, which means our predictions contain more high-quality examples. \tab{\ref{tab:averageprecision}} reveals the average precision of 2D and 3D object detection. It is clear that our VS3D has superior performance over the compared baselines. For example, the AP\textsubscript{2D} of VS3D is over $50\%$ higher than the baselines under IoU threshold 0.3 and 0.5. Three baseline approaches utilize selective search~\cite{UijlingsIJCV2013} to generate object proposals, which has been a prevailing unsupervised object proposal approach. However, the rich 3D geometric information is left unexplored in these methods. The baseline approaches can be improved using our object proposal methods as is shown in \tab{\ref{tab:apwithupm}}. We also compare our weakly supervised VS3D with fully supervised methods in \tab{\ref{tab:gapweakfull}}. Results are obtained on the public KITTI~\cite{geiger2012kitti} validation set widely used by previous work~\cite{chen2016monocular, chen20153dop, chen2017multiview}.

An interesting phenomenon could be observed by comparing our VS3D with different input data types. Generally speaking, if the evaluation metric is in 3D and the IoU requirement is high, the LiDAR based version would be at an advantage. But for 2D metrics such as 2D recall and AP\textsubscript{2D}, as well as 3D metrics with low IoU threshold, the monocular and stereo versions could have a better performance. This phenomenon could be interpreted as follows. For a 3D metric with high IoU thresholds, the requirement of 3D localization could be far higher, and LiDAR is good at providing such a geometric precision. The point clouds generated by monocular and stereo images cannot reach the precision as a LiDAR does. On the contrary, for a 2D metric or a 3D metric with low IoU thresholds, the requirement of 3D localization is much lower. The point clouds generated from images have a higher resolution that LiDAR point clouds and are more suitable for semantic scene understanding, which make it possible for image-based approaches to have better performance. It should be mentioned that if the dataset contains more night-time scenes, the results of image-based versions could be dropped behind the LiDAR based version. Most RGB cameras are passive sensors and are affected by darkness, while LiDARs are active sensors with built-in light sources and thus are less influenced by the external illumination. Therefore, the optimal approach should be able to appropriately combine cameras and LiDARs, which can complement each other in different scenarios.

In addition to the quantitative results, we also present the qualitative results in \fig{\ref{fig:qualitative}}. The three columns correspond to the Mono, Stereo and LiDAR version of VS3D respectively. It is shown that for objects faraway from the camera origin, the LiDAR version works better than the Mono and Stereo versions, which is reasonable because the quality of point cloud generated from images decreases as the distance increases. In the LiDAR version, predicted 3D bounding boxes are aligned with the ground truth boxes, which is hard to achieve when the ground truth are not available in training.

\subsection{Ablation study on UPM}

The proposed unsupervised 3D object proposal module (UPM) selects and align predefined anchors with high objectiveness confidence, removing over 98\% of the total anchors that are redundant. Our UPM is based on the normalized point cloud density (NPCD) that is a distance-invariant indicator of the presence of objects. In order to validate the effectiveness of our approach, we replace NPCD with another two strategies and compare the bounding box recall rate. The first is an inclusive strategy (INC), where the predefined anchors are all kept without being filtered. The second is based on point cloud density (PCD), where the PCD is measured without the proposed normalization step. Results are shown in \fig{\ref{fig:recall_curve}}. It is clear that our NPCD demonstrates better performance than INC and PCD. The gap between NPCD and PCD is mainly due to the normalization step. PCD can reflect the objective confidence in an object proposal, but is severely influenced by distance. Most of the distant anchors are filtered because they have a low point cloud density, even if they contain objects. Therefore, objects far away will be missing in the proposals unless the distance interference is removed, which is achieved in NPCD.

\section{Conclusion}
This paper presents a pioneering work on weakly supervised learning of 3D object detection from point clouds. Our pipeline consists of the unsupervised 3D object proposal module (UPM) and the cross-modal transfer learning module. UPM takes the raw point cloud as input and outputs the 3D object proposals. Without ground truth supervision, UPM leverages the normalized point cloud density to identify the 3D anchors potentially containing objects. Object proposals predicted by UPM are classified and refined by the student network to produce the final detection results. The point cloud based student network is trained by an image based teacher network via transferring the knowledge from existing image datasets to the point cloud domain. Comprehensive experiments demonstrate our promising performance in diverse evaluation settings. Our method can potentially reduce the need of manual annotation and facilitate the deployment of 3D object detectors in new scenarios.

\bibliographystyle{ACM-Reference-Format}
\bibliography{reference}

\end{document}